\documentclass[conference]{IEEEtran}
\usepackage{cite}
\usepackage{amsmath,amssymb,amsfonts}
\usepackage{algorithmic}
\usepackage{graphicx}
\usepackage{textcomp}
\usepackage{xcolor}
\usepackage{multirow}
\usepackage{footnote}
\usepackage{hyperref}
\usepackage{mathtools}
\def\BibTeX{{\rm B\kern-.05em{\sc i\kern-.025em b}\kern-.08em
    T\kern-.1667em\lower.7ex\hbox{E}\kern-.125emX}}
\begin{document}

\title{Attention-augmented Spatio-Temporal Segmentation for Land Cover Mapping}

\makeatletter
\newcommand{\linebreakand}{%
  \end{@IEEEauthorhalign}
  \hfill\mbox{}\par
  \mbox{}\hfill\begin{@IEEEauthorhalign}
}
\makeatother

\author{
    \IEEEauthorblockN{Rahul Ghosh\IEEEauthorrefmark{1} }
    \IEEEauthorblockA{
        \textit{University of Minnesota}\\
        Minneapolis, US \\
        ghosh128@umn.edu
    }
    \and
    \IEEEauthorblockN{Praveen Ravirathinam\IEEEauthorrefmark{1}}
    \IEEEauthorblockA{
        \textit{University of Minnesota}\\
        Minneapolis, US \\
        pravirat@umn.edu
    }
    \and
    \IEEEauthorblockN{Xiaowei Jia}
    \IEEEauthorblockA{
        \textit{University of Pittsburgh}\\
        Pittsburgh, US \\
        xiaowei@pitt.edu
    }
    \linebreakand
    \IEEEauthorblockN{Chenxi Lin}
    \IEEEauthorblockA{
        \textit{University of Minnesota}\\
        Minneapolis, US \\
        lin00370@umn.edu
    }
    \and
    \IEEEauthorblockN{Zhenong Jin}
    \IEEEauthorblockA{
        \textit{University of Minnesota}\\
        Minneapolis, US \\
        jinzn@umn.edu
    }
    \and
    \IEEEauthorblockN{Vipin Kumar}
    \IEEEauthorblockA{
        \textit{University of Minnesota}\\
        Minneapolis, US \\
        kumar001@umn.edu
    }
    \linebreakand
    \IEEEauthorblockA{\IEEEauthorrefmark{1}authors contributed equally to this research}  

}

\maketitle

\begin{abstract}
The availability of massive earth observing satellite data provides huge opportunities for land use and land cover mapping. However, such mapping effort is challenging due to the existence of various land cover classes, noisy data, and the lack of proper labels. Also, each land cover class typically has its own unique temporal pattern and can be identified only during certain periods. In this article, we introduce a novel architecture that incorporates the UNet structure with Bidirectional LSTM and Attention mechanism to jointly exploit the spatial and temporal nature of satellite data and to better identify the unique temporal patterns of each land cover class. We compare our method with other state-of-the-art methods both quantitatively and qualitatively on two real-world datasets which involve multiple land cover classes. We also visualise the attention weights to study its effectiveness in mitigating noise and in identifying discriminative time periods of different classes. The code and dataset used in this work are made publicly available for reproducibility.
\end{abstract}


\section{Introduction}
\label{Sec:Introduction}
Growth in the world's population and the acceleration of urbanization are straining already scarce natural resources and food supplies, which must scale up to keep pace with growing demand. The consequences of the resulting large-scale changes include tremendous stresses on the environment, as well as challenges to our ability to feed the world's population, that could be calamitous at the current rate of change if they are not managed sustainably. Timely information on land use and land cover changes can provide critical information at desired spatial and temporal scales to assist in decision making for development investment and sustainable resource management. In particular, mapping crops is a key step towards many applications, such as forecasting yield, guiding sustainable management practices and evaluating progress in conservation efforts.

\begin{figure}[t]
    \centering
    \includegraphics[width = \linewidth]{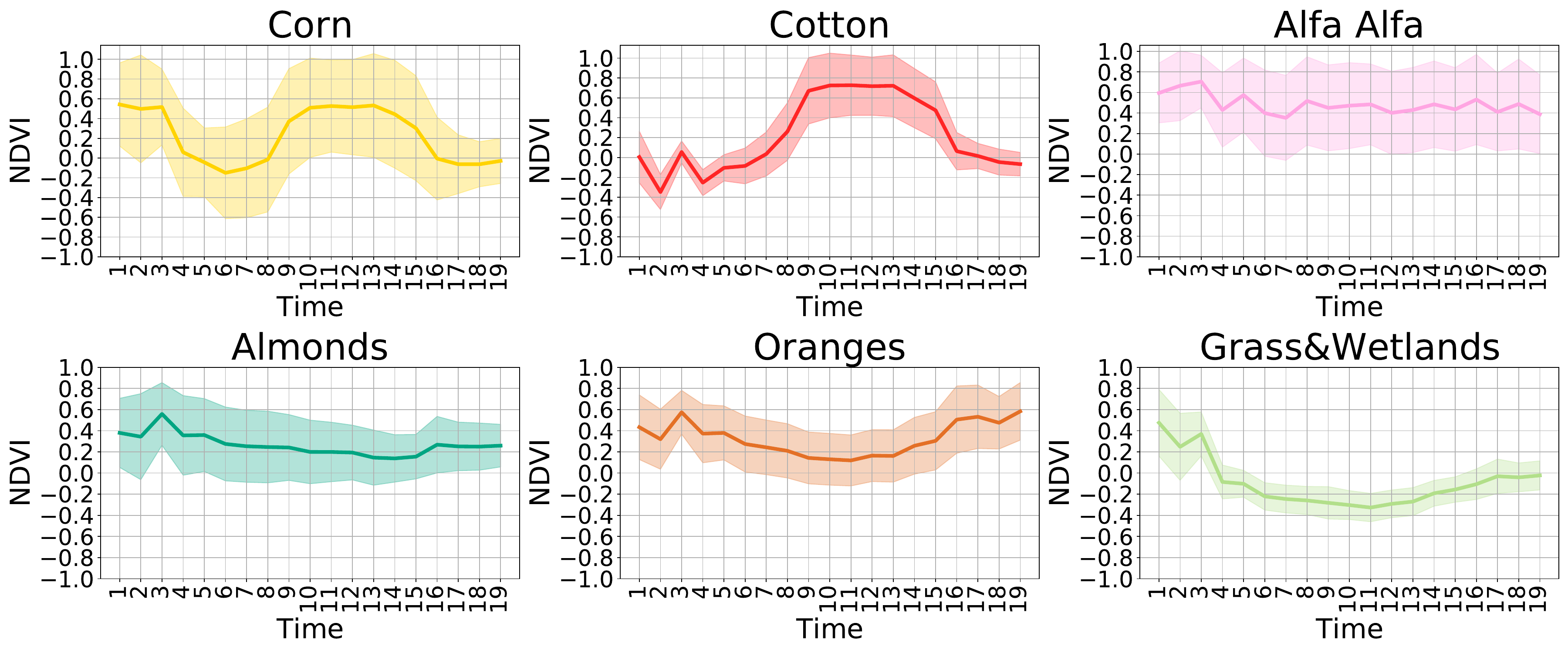}
    \caption{Yearly Normalized Difference Vegetation Index (NDVI) time series for some land cover classes.}
    \label{fig:ndvi}
\end{figure}

Recent advances in storing and processing remote sensing data collected by sensors on-board aircrafts or satellites provide tremendous potential for mapping a variety of land covers. Deep learning on remote sensing has shown promising results  for mapping specific land covers, e.g.,  plantations~\cite{jia2017incremental}, agricultural facilities~\cite{handan2019deep}, roads~\cite{xie2015transfer}, and buildings~\cite{nayak2020semi}, by extracting their distinct spatial and temporal patterns. In particular, Convolutional Neural Networks (CNN)-based models have been widely used in  automated  land  cover  mapping given their capacity to capture the spatial correlation, i.e., land covers are often contiguous over space~\cite{maggiori2016convolutional,postadjian2017investigating}. Extracting the spatial information also assists in mitigating the noise of remote sensing data  at individual locations (e.g., aerosols). While CNNs have mostly been used for studying land covers from a single image, Recurrent Neural Networks (RNN) and Long-Short Term Memory (LSTM) can take advantage of temporal patterns~\cite{jia2014land,russwurm2017temporal} obtained from multi-temporal remote sensing data. Many land covers, e.g., crops, are indistinguishable at a single time step but require the modeling of their growing or seasonal patterns.  For e.g. certain crops may look similar to barren lands or wetlands after they are harvested but look different in growing season. Thus, successful detection of these land covers depends on the extraction of their distinctive temporal patterns. 
Although these architectures model the temporal information, majority of them do not consider the spatial correlation shown by land covers.

Researchers have also built spatio-temporal models for land cover mapping~\cite{ji20183d,jia2017incremental,mazzia2020improvement} and these approaches have shown encouraging results in isolated scenarios for studying a specific land cover. However, there is a major challenge that prevents these approaches from successfully mapping crops in large regions. In particular, these approaches use CNN and RNN in a straightforward way without fully exploiting the characteristics of land covers. Different crops show discriminative signatures at different points of time~\cite{jia2019spatial,sakamoto2010detecting}. Moreover, different crops have different seeding time and harvesting time,  depending on the weather conditions. To highlight these characteristics, we show the averaged yearly vegetation index for different crops in Fig.~\ref{fig:ndvi}. In addition, the images from some time steps may be affected by natural disturbances (e.g. clouds, weather) or data acquisition errors which can severely degrade the classification performance. Thus, identifying these discriminative periods automatically for different locations and different years and filtering out the noisy time steps is essential for distinguishing between different types of crops accurately.

In this paper, we propose Spatio-Temporal segmentation networks with ATTention (STATT) for automated land-cover detection. Specifically, the proposed model uses an UNet~\cite{ronneberger2015u} encoder to embed the spatial information from each image into a more representative feature space, and then introduces a Bidirectional Long-Short Term Memory (Bi-LSTM) layer to capture long-term temporal dependencies and generate spatio-temporal features. We further use the attention networks to aggregate the obtained spatio-temporal features at different time steps. The attention networks help identify the most relevant time steps for classifying  land covers and mitigate the impact from noisy images by assigning lower weights to those time steps. 

We further enhance the detection by fusing spatio-temporal features extracted at multiple resolutions through the skip connections. Specifically, for each resolution level (obtained through the UNet encoder) we use the same weights obtained from the aforementioned attention network to aggregate the spatial features extracted by the encoder at multiple time steps. These aggregated features at different resolutions are transferred to the UNet decoder to improve the segmentation.

We show the superiority of our method over existing spatio-temporal learning methods in mapping crops in different regions, including agriculture-intensive areas in US and another region that is dominated by tree plantations in Africa. We also showcase the usability of attention weights in filtering out noisy time steps and also identifying time steps in which the classes are distinguishable.

Our contributions can be summarized as follows:
\begin{itemize}
    \item We develop a spatio-temporal segmentation pipeline that leverages the data available in the form of temporal satellite imagery.
    \item We augment the spatio-temporal segmentation pipeline using an attention network to identify discriminative time-periods and reduce the effect of atmospheric noises such as clouds.
    \item We release the code and dataset used in this work to promote reproducibility (Please refer to \url{https://drive.google.com/drive/folders/1CSHjWgXlLx3BF-LFRDE0JKTLo-Dqhd9M?usp=sharing}).
\end{itemize}


\section{Related Work}
\label{Sec:Related Work}

\paragraph{Land Use and Land Cover mapping}
In recent literature, machine learning techniques, especially deep learning (DL), have been heavily used for land use and land cover (LULC) mapping. In particular CNNs \cite{hu2018deep,stoian2019land} have been used to extract representations for both spectral and spatial information, whereas RNN and LSTM \cite{jia2017incremental} make use of the temporal information in modeling land cover transitions and have shown promising performance in sequence labelling. Land Cover mapping can also be framed as a semantic segmentation problem \cite{ulmas2020segmentation,su2019land,saralioglu2020semantic}, where each pixel in an aerial/satellite image must classified into one of several land cover classes.

\paragraph{Spatio-temporal modeling for LULC changes}
To capture both the spatial and temporal contextual, existing works commonly follow the end-to-end learning paradigm using a combination of convolutional and recurrent networks. Shi et al. \cite{xingjian2015convolutional} proposed a novel ConvLSTM layer which replaced the standard feed-forward neural network in an LSTM with a convolutional one. Ru{\ss}wurm et al. \cite{russwurm2018convolutional} applied a single ConvLSTM layer to detect cloud obstruction. Another approach~\cite{garnot2019time} used a shared CNN to embed the image time-series and the resulting sequence was fed to an RNN for spatio-temporal image classification. This approach is inspired by the LRCN model~\cite{donahue2015long} which has been successfully used for activity recognition in videos. All these methods have been focused on the image classification and do not address the semantic segmentation problem. 3D convolutions is another method that has been used in land cover classification \cite{ji20183d} to capture the temporal information of satellite images. The convolutional approach suffers from the inability to handle sequences of different lengths.

\begin{figure*}[t]
    \centering
    \includegraphics[width = \textwidth]{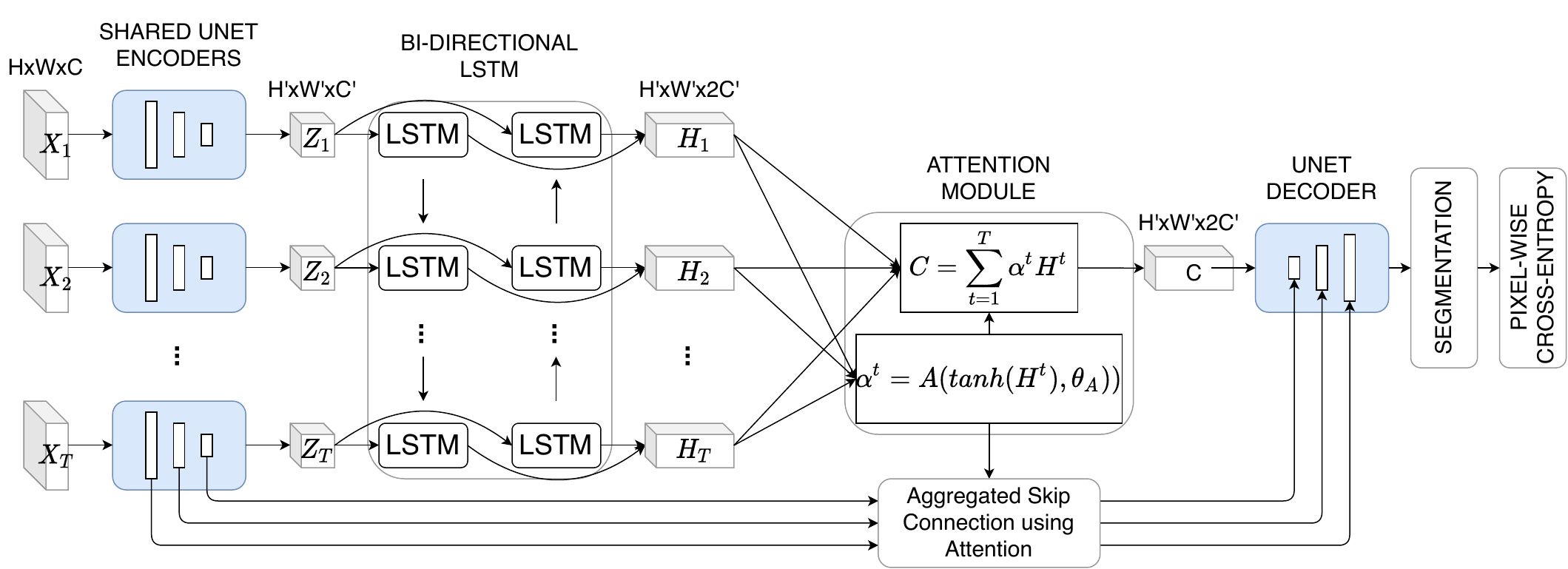}
    \caption{A diagrammatic representation of STATT.}
    \label{fig:architecture}
\end{figure*}

\paragraph{Semantic Segmentation}
One of the most fundamental techniques used in semantic segmentation is the Fully Convoluted Network (FCN)~\cite{long2015fully} which supplements the output of the deeper layers with that of the shallower layers to increase the resolution of the prediction. Researchers have also developed several variants of FCN such as SegNet \cite{badrinarayanan2017segnet}, DeconvNet \cite{noh2015learning} and UNet \cite{ronneberger2015u}. Given the effectiveness of UNet in a variety of segmentation tasks,  we adopt UNet as the base of our proposed method. UNet consists of two paths, contraction path (encoder) and symmetric expanding path (decoder). The encoder consists of a stacked set of convolutional and max-pooling layers, that captures the context and a semantic understanding of the image. On the other hand, the decoder consists of convolutional and upconvolutional layers, tasked with generating precise label maps from the output of the encoder.


\section{Problem definition and preliminaries}
\label{Sec:Problem Setting}

\subsection{Problem Setting}
In this paper, we consider the task of land cover mapping and frame it as a semantic segmentation problem using the multi-spectral satellite/aerial image time-series. In particular, we aim to predict the land cover class $\boldsymbol{l} \in \{1,...,\text{L}\}$ for each pixel in an image. During the training process, we have access to image time-series and corresponding labels, which can be described as follows:
\begin{itemize}
    \item Input image time-series $X = [X^1, \dots, X^T]$, where each $X^t \in \mathbb{R}^{\text{H}\times \text{W}\times \text{C}}$ is an aerial/satellite image of size $(\text{H},\text{W})$ at time $t$ with $\text{C}$ multi-spectral channels.
    \item Labels $Y \in \mathbb{\{0,1\}}^{\text{H}\times \text{W}\times \text{L}}$ in one-hot representation, where $\text{L}$ is the number of land-cover classes.
\end{itemize}

\subsection{Segmentation network}
A segmentation network $f(X_i;\theta)$ aims to predict the label of each pixel for an image $X_i$. The parameter $\theta$ is estimated through a training process on a labeled dataset by minimizing an objective function of empirical risk, such as the pixel-wise cross entropy, as follows:

\begin{equation}
    \label{eq:CrossEntropy}
    \mathcal{L}_{CE}(\boldsymbol{\theta}|\mathbf{X},\mathbf{Y}) = -\frac{1}{\text{NHW}}\sum_i\sum_{(h,w)}\sum_{k} (Y_i)_{h,w}^k\log f(X_i;\theta)_{h,w}^k
\end{equation}

where $f(X_i;\theta)_{h,w}^k$ is the likelihood of the $(h,w)$’th pixel belonging to the class $k$ as predicted by the fully-convolutional network and $(Y_i)_{h,w}^k = 1$ if the $(h,w)$’th pixel of the image $i$ belongs to the class $k$.

\section{Method}
\label{Sec:Method}
Traditional segmentation architectures like UNET \cite{ronneberger2015u} are designed for single image semantic segmentation and thus fail to capture the temporal information present in the satellite/aerial image time-series. In this section, we describe the proposed STATT method for land cover mapping (see Fig.~\ref{fig:architecture}). Our method extends the UNet architecture~\cite{ronneberger2015u} to capture the spatial and temporal land cover patterns, identify discriminative land cover patterns and noisy time periods, and fuse multi-resolution information through skip connections.  

\subsection{Spatio-Temporal modeling}
The standard UNet encodes the semantic information of a single image into a hidden representation, which is then decoded to produce the segmentation results along with the help of skip connections to recover fine-level details. Similarly, in our proposed architecture, each satellite image $X^t \in \mathbb{R}^{\text{H}\times \text{W}\times \text{C}}$ of a sequence $X = [X^1, \dots, X^T]$ is passed through an convolutional encoder $E(\,\cdot\,;\theta_E)$ to encode the spatial information of each image into a latent representation $Z^t \in \mathbb{R}^{\text{H}'\times \text{W}'\times \text{C}'}$. This results in a sequence of $(\text{H}',\text{W}')$ spatial features of dimension $\text{C}'$ which is represented as $Z = [Z^1, \dots, Z^T]$ in dimension $\text{T}\times \text{H}'\times \text{W}'\times \text{C}'$. This ordered set of spatial features is then passed into a recurrent sequence learning module to capture the temporal dependencies in the time series of spatial features. Specifically, given a sequence of input features $Z_{ij} = [Z^1_{ij}, \dots, Z^T_{ij}]$, we use an LSTM network to compute a sequence of hidden states $H_{ij} = [H^1_{ij}, \dots, H^T_{ij}]$. Each hidden representation $H_{ij}^t$ is generated using the hidden state $H_{ij}^{t-1}$ and cell state $C_{ij}^{t-1}$ from the previous time step and the input spatial-features using the following set of equations:

\begin{align}
    \begin{rcases}
        F_{ij}^t &= \sigma (W_H^FH_{ij}^{t-1} + W_Z^FZ_{ij}^{t})\\
        I_{ij}^t &= \sigma (W_H^IH_{ij}^{t-1} + W_Z^IZ_{ij}^{t})\\
        O_{ij}^t &= \sigma (W_H^OH_{ij}^{t-1} + W_Z^OZ_{ij}^{t})\\
        G_{ij}^t &= \text{tanh}(W_H^GH_{ij}^{t-1} + W_Z^GZ_{ij}^{t})\\
        C_{ij}^t &= F_{ij}^t\odot C_{ij}^{t-1} + I_{ij}^t\odot G_{ij}^t\\
        H_{ij}^t &= O_{ij}^t \odot \text{tanh}(C_{ij}^t)
    \end{rcases} {i,j}\in \left(\text{H}', \text{W}'\right)
    \label{eq:Bi-LSTM}
\end{align}

Different crops have specific growing and harvesting patterns, and thus, capturing their specific growing and harvesting patterns are essential in successful identification. Traditional LSTMs tend to be biased towards the information provided by the previous time step. To avoid the previous time step bias and learn spatio-temporal features at each time step using sufficient context of growing and harvesting patterns we use a Bi-LSTM. Specifically, we build two LSTM structures called the forward LSTM and backward LSTM using eq \ref{eq:Bi-LSTM}. The two LSTM structures are the same except that the time-series is reversed for the backward LSTM. The forward LSTM models the growing patterns whereas the backward LSTM models the harvesting patterns of a crop. The spatio-temporal features at each time step are obtained by concatenating the hidden representation of both the forward and backward LSTMs as shown in figure \ref{fig:architecture}. The obtained hidden representations $H = [H^1, \dots, H^T]$ capture  the spatial information as well as the temporal information by modeling the change in the spatial features.

\subsection{Attention based aggregation}
The spatio-temporal hidden representations are aggregated using an attention mechanism which calculates the relevance scores for each time-step based on their contribution to the classification performance. Specifically, we use a feed-forward neural network $f_{\theta_A}(\cdot)$ with parameters $\theta_A$, followed by a spatial averaging and softmax normalization over all the time steps, as follows:
\begin{align}
    \label{eq:AttentionWeights}
    \alpha^t &= \text{softmax}(\frac{1}{\text{H}'\times \text{W}'}\sum_{\left(i,j\right) \in \left(0,0\right)}^{\left(\text{H}',\text{W}'\right)}f_{\theta_A}(H_{ij}^t))
\end{align}

Using the obtained attention weights, we combine the spatio-temporal features $H^t$ of all the time steps into an aggregated spatio-temporal features $C$, as follows:

\begin{align}
    \label{eq:AttentionWeights_agg}
    C &= \sum_{t=1}^{T} \alpha^t H^t
\end{align}

Here the higher value of  $\alpha^t\in[0,1]$ indicates that the time step $t$ contains critical information for detecting target land covers. In contrast, the lower value of  $\alpha^t$ indicates less importance of the time step $t$, either because this time step contains much noise or it is out of the discrimiantive period.

\subsection{Aggregating multi-resolution features through skip connections}
Given the weighted spatio-temporal features $C$, we build a convolutional decoder $D(\,\cdot\,,\theta_D)$ to generate segmentation labels.  Additionally, we use skip connections as a direct pathway to fuse spatio-temporal features extracted by the encoder at multiple resolutions. Specifically, our architecture has multiple convolution blocks in the encoder and multiple up-convolution blocks in the decoder. Here these blocks extract spatial features at different spatial resolutions. For the $k^{th}$ convolution block in the encoder, it  outputs multiple intermediate images $(Z_1^k, \dots, Z^k_T)$ over multiple time steps. We use the same attention weights calculated in Eq.~\ref{eq:AttentionWeights_agg} and aggregate the outputs $Z^k_t$ using Eq.~\ref{eq:AttentionAggregation}, as follows:

\begin{equation}
    \label{eq:AttentionAggregation}
    C_k = \sum_{t=1}^{T} \alpha^t Z^k_t
\end{equation}

The obtained spatio-temporal features at the specific resolution (corresponding to $k^{th}$ block) are then merged to the corresponding layer in the decoder with the same resolution. The output of the decoder $D(C,\theta_D)$ is passed through a linear classification layer $f(,\theta_c)$ followed by a softmax function to generate class probabilities. This model can be trained using a pixel-wise cross entropy loss as shown in Eq.~\ref{eq:CrossEntropy}.

\section{Experimental Results}
\label{Sec:Evaluation}

\subsection{Datasets and Implementation details}
We evaluate our proposed strategy for spatio-temporal semantic segmentation on two real-world applications of great societal impacts. In the first example, we investigate crop mapping in the California, US which has over 30 classes of crops and vegetables. In the second example, we aim to map cashew plantation in Benin, which contribute nearly 10\% of the country's export income. Accurate cashew plantation mapping provides inventory information of cashew to the Benin government to assist the distribution of their recent \$100 million loan from World Bank, aiming at further developing the cashew industry. Mapping crops is a key step towards many applications, such as forecasting yield, guiding sustainable management practices and evaluating progress in conservation efforts.

\paragraph{D1:} \textbf{Sentinel based crop mapping for Central Valley in California, USA}
In this dataset, we investigate crop mapping in the US Midwest, the world's bread basket. This data set contains multi-spectral images observed by the Sentinel-2 Constellation for the year 2018 (see the data release link in Section~\ref{Sec:Introduction}). The Sentinel-2 data product has 13 spectral bands \footnote{\url{https://developers.google.com/earth-engine/datasets/catalog/COPERNICUS_S2_SR##bands}} at three different spatial resolutions of 10, 20 and 60 metres. We leave out the atmospheric bands (Band 1, 9 and 10) of 60 metres resolution and re-sample all the bands to 10 metres using the nearest neighbour method. The AOI corresponds to the Sentinel tile T11SKA and has a wide variety of crop types, resulting in a challenging task for land cover mapping. We use images for every 15 days from January to December (in total 24 time steps) with the size of (10980, 10980). This dataset contains a mixture of clean and noisy (cloudy) images. For our experiment, we aim to classify each pixel to a class label $l \in \{$ \texttt{corn, cotton, wheat, alfalfa, tomatoes, grapes, tree crops, almonds, walnut, pistachio, grass \& wetlands, barren land/idle land, open water and urban} $\}$., where the labels for this data set are taken from the Cropland Data Layer (CDL) \footnote{\url{https://nassgeodata.gmu.edu/CropScape/}} provided by United States Department of Agriculture (USDA) as the ground truth. The CDL layer however is at 30m resolution and STATT predicts at 10m resolution. To overcome this we resample the CDL layer to 10m resolution as well, but this potentially leads to inaccurate and noisy labels at the spatial boundaries between two classes. To reduce the effect of noisy labels from CDL caused by resampling, we use a label cleaning strategy where for each class label we do a 1 pixel erosion followed by removal of connected components of size less than 10. This removes boundary pixels and also isolated pixels of a class. We replace the removed pixels with a new called unknown and do not consider these pixels while training and testing. 

\paragraph{D2:} \textbf{Planet based Cashew tree-crop mapping in Benin, Africa}
In this dataset, we aim to map cashew plantation in Benin, which contribute nearly 10\% of the country's export income. Benin government is actively looking for inventory information of cashew to assist the distribution of their recent \$100 million loan from World Bank, aiming at further developing the cashew industry. This data set contains multi-spectral monthly composites from Planet Labs~\footnote{\url{https://www.planet.com/products/basemap/}} for a region in Benin, Africa, where cashew tree crops are a major source of income for farmers (they contribute nearly 10\% of the countries export income). The images are of size (2700, 2400) pixels, having 4 spectral bands namely red, green, blue and NIR(near infrared) at a spatial resolution of 3 metres. For our experiment, we aim to classify each pixel into a class label $l \in \{$ Forest, Barren land, Cashew, Urban $\}$. The ground truth was created using manual annotation over the entire study region using high resolution Airbus imagery provided by our collaborators in Benin, Africa~\footnote{Given the proprietary nature of the Planet Lab composite and the Airbus imagery, we do not have permission to make this data publicly available.}.

In both datasets D1 and D2, we divide the whole image into $10\times10$ grids and randomly select 60\%, 20\%, 20\% grids for training, validation and testing, respectively. For D1 we use an input patch size of $32\times32$ to output a patch of size $16\times16$ while in D2, we use an input patch size of $64\times64$ to output a patch of size $60\times60$. The same settings were used for all baseline experiments and all the methods were trained using the Adam optimizer with a batch-size of 32.

\subsection{Architecture details}
Due to the different spatial resolutions of the datasets D1 and D2 we found that varying the architecture of STATT between the two datasets is beneficial. In this section, we present the details of the architecture used for the datasets.

\paragraph{STATT for D1:} \textbf{Sentinel based crop mapping for Central Valley in California, USA}
For this task, we use three convolutional blocks in our encoder each having two convolutional layers. Thus there are six convolutional layers having ${64, 64, 128, 128, 256, 256}$ channels and filters of size $3\times3$. To downsample the the output of the convolutional blocks we use max-pooling of size $2\times2$ after the first and second convolutional blocks. In the decoder, we have two convolutional blocks each of which consists two convolutional layers. The four convolutional layers of the decoder have ${128, 128, 64, 64}$ channels respectively. To upsample the output we add transposed-convolutional layers before the first and second convolutional block of the decoder having ${128, 64}$ channels respectively and kernel size of $2\times2$. Finally we add a fully-connected layer with input dimension of 64 and output dimension equal to the number of classes i.e. 14.

\paragraph{STATT for D2:} \textbf{Planet based Cashew tree-crop mapping in Benin, Africa}
Since this dataset has a higher spatial resolution, we use larger input patches for sufficient spatial context and a deeper architecture for effective capturing of spatial information. Specifically, we use five convolutional blocks in our encoder each having two convolutional layers. Thus there are 10 convolutional layers having ${16, 16, 32, 32, 64, 64, 128, 128, 256, 256}$ channels and filters of size $3\times3$. To downsample the the output of the convolutional blocks we use max-pooling of size $2\times2$ after each of the last but one convolutional blocks. In the decoder, we have four convolutional blocks each of which consists two convolutional layers. The eight convolutional layers of the decoder have ${128, 128, 64, 64, 32, 32, 16, 16}$ channels respectively. To upsample the output we add transposed-convolutional layers before the first and second convolutional block of the decoder having ${128, 64, 32, 16}$ channels respectively and kernel size of $2\times2$. Finally we add a fully-connected layer with input dimension of 16 and output dimension equal to the number of classes i.e. 4.

In the above architectures, we use a one-layer bidirectional LSTM network with 256 hidden units at the bottleneck.

\begin{table*}[t]
    \centering
    \caption{Comparison with baselines in terms of Crop Wise as well as Mean F1 Score. The numbers in bold and with $*$ symbol correspond to the best and second best method for each row respectively. We also show the time taken for training(for 1 epoch) and testing of baselines and our method on D1.}
    \resizebox{\textwidth}{!}{%
        \begin{tabular}{|c|c|c||ccccc||cc|}
            \hline
                                & \hfill CLASS(Percentage)              & \hfill Count      & UNet              & Bi-LSTM Attn      & CALD      & ConvLSTM          & 3D-CNN            & STMean            & STATT\\\hline\hline
            \multirow{15}*{D1}  & \hfill Barren Land (13.77)            & \hfill 2139439    & 0.6913            & 0.4662            & 0.6571    & 0.7486            & 0.6860            & 0.7498*           & \textbf{0.7761}\\
                                & \hfill Urban (11.97)                  & \hfill 1859910    & 0.8282            & 0.5869            & 0.6918    & 0.8110            & 0.8143            & \textbf{0.8705}   & 0.8690*\\
                                & \hfill Grapes (11.58)                 & \hfill 1799302    & 0.8013            & 0.7959            & 0.7103    & 0.7441            & 0.8359            & 0.8477*           & \textbf{0.8493}\\
                                & \hfill Almonds (10.79)                & \hfill 1676985    & 0.8357            & 0.7684            & 0.7640    & 0.8407*           & 0.8389            & 0.8283            & \textbf{0.8426}\\
                                & \hfill Grass (10.22)                  & \hfill 1588288    & 0.7859            & 0.7315            & 0.7582    & 0.7862            & 0.7595            & 0.8124*           & \textbf{0.8290}\\
                                & \hfill Tree crops (09.77)             & \hfill 1517775    & 0.7770            & 0.7137            & 0.6443    & 0.7118            & 0.7952            & 0.8064*           & \textbf{0.8249}\\
                                & \hfill Corn (07.73)                   & \hfill 1200614    & 0.8172            & 0.9223            & 0.9105    & \textbf{0.9332}   & 0.9217            & 0.9220            & 0.9262*\\
                                & \hfill Cotton (07.04)                 & \hfill 1093876    & 0.9335            & 0.9519            & 0.9575    & 0.9623            & \textbf{0.9696}   & 0.9586            & 0.9676*\\
                                & \hfill Pistachio (05.17)              & \hfill 802644     & 0.6702            & 0.5821            & 0.5875    & 0.6808            & 0.8202*           & 0.7921            & \textbf{0.8358}\\
                                & \hfill Alfa Alfa (03.87)              & \hfill 600932     & 0.8438            & 0.7971            & 0.6148    & 0.8098            & 0.8706            & \textbf{0.8907}   & 0.8798*\\
                                & \hfill Tomatoes (03.07)               & \hfill 477757     & 0.7046            & 0.8891            & 0.9069    & 0.9244*           & 0.9149            & 0.8803            & \textbf{0.9406}\\
                                & \hfill Walnut (02.42)                 & \hfill 376654     & 0.8457            & 0.5984            & 0.4758    & 0.7511            & \textbf{0.8666}   & 0.8456            & 0.8477*\\
                                & \hfill Wheat (02.20)                  & \hfill 342596     & 0.3228            & \textbf{0.7522}   & 0.6594    & 0.7369            & 0.7397            & 0.6887            & 0.7431*\\
                                & \hfill Water (0.40)                   & \hfill 62738      & \textbf{0.6967}   & 0.5566            & 0.5649    & 0.5700            & 0.5650            & 0.6533            & 0.6756*\\\hline
                                & \hfill MEAN                           & \hfill 15539510   & 0.7539            & 0.7223            & 0.7074    & 0.7865            & 0.8142            & 0.8247*           & \textbf{0.8434}\\ \hline
            \multirow{5}*{D2}   & \hfill Tree crops (51.63)             & \hfill 2071563    & 0.8264            & 0.8180            & 0.8197    & 0.8864            & 0.8751            & 0.8969*           & \textbf{0.9103}\\
                                & \hfill Barren Land (30.75)            & \hfill 1233691    & 0.7723*           & 0.6742            & 0.6521    & 0.7362            & 0.7263            & 0.7337            & \textbf{0.7832}\\
                                & \hfill Cashew (17.10)                 & \hfill 685997     & 0.6782            & 0.7817            & 0.7572    & 0.7865            & 0.7929            & 0.8019*           & \textbf{0.8410}\\
                                & \hfill Urban (0.53)                   & \hfill 21304      & 0.5513            & 0.4395            & 0.4349    & 0.4807            & 0.4555            & 0.5586*           & \textbf{0.5857}\\ \hline
                                & \hfill MEAN                           & \hfill 4012555    & 0.7071            & 0.6716            & 0.6660    & 0.7225            & 0.7124            & 0.7478*           & \textbf{0.7800}\\ \hline \hline
                                & \multicolumn{2}{r||}{TRAIN TIME(in sec/epoch)}            & 11040             & 517               & 4668      & 3833              & 1812              & 1141              & 1154\\
                                & \multicolumn{2}{r||}{TEST TIME(in sec)}                   & 4008              & 176               & 1848      & 546               & 321               & 241               & 245 \\ \hline
        \end{tabular}
    }
    \label{Tab:tableResults}
\end{table*}

\subsection{Baselines}
We compare the performance of our method against the following baselines:
\begin{enumerate}
    \item \textbf{UNet} We use a separate single-image UNet architecture $f_{UNet}^i(X^i), i\in(1, \dots, T)$ to model the spatial information present in each multi-spectral image time step. The final prediction of each pixel is obtained by taking the softmax on the average of the models outputs, i.e., $y = \text{softmax}(\frac{\sum_{i\in(1, \dots, T)}f_{UNet}^i(X^i)}{T})$. This baseline models the spatial information provided by each image in the time-series and lacks as it doesn't model the temporal information.
    \item \textbf{Bi-LSTM Attn} We use a Bi-directional LSTM model with attention followed by a linear classifier on image pixel time-series. This baseline don't model the spatial and temporal information jointly and uses the spectral and temporal information while disregarding the spatial information of the images
    \item \textbf{CALD} \cite{jia2019spatial} This baseline method provides one such solution for joint modeling of the spatio-temporal information, where the authors use a context-aware LSTM model which also considers the information provided by the neighborhood pixels for predicting the class labels of the pixel of interest.
    \item \textbf{3D-CNN}~\cite{ji20183d} We adapt the UNet architecture and replace the 2D convolution layers of the encoder with 3D CNNs. We use the same configuration of filter size as used by the authors in their evaluation. The skip connections in the UNet are replaced by an average skip connection of the multi-temporal outputs. 3D-Convolutions only consider $L$ temporally adjacent frames (e.g. $L=3$) and this problem is addressed to an extent by temporal pooling. However, this does not allow the modeling of temporal evolution of crops.
    \item \textbf{ConvLSTM}~\cite{xingjian2015convolutional} We adapt the UNet architecture and replace the convolution layers of the encoder with the ConvLSTM layers proposed by the authors. Since this methods used an LSTM network at each convolution layer, this results in a model with a significantly large number of parameters making training and testing slow when compared to other baselines.
\end{enumerate}

\begin{figure}[t]
    \centering
    \includegraphics[width=\linewidth]{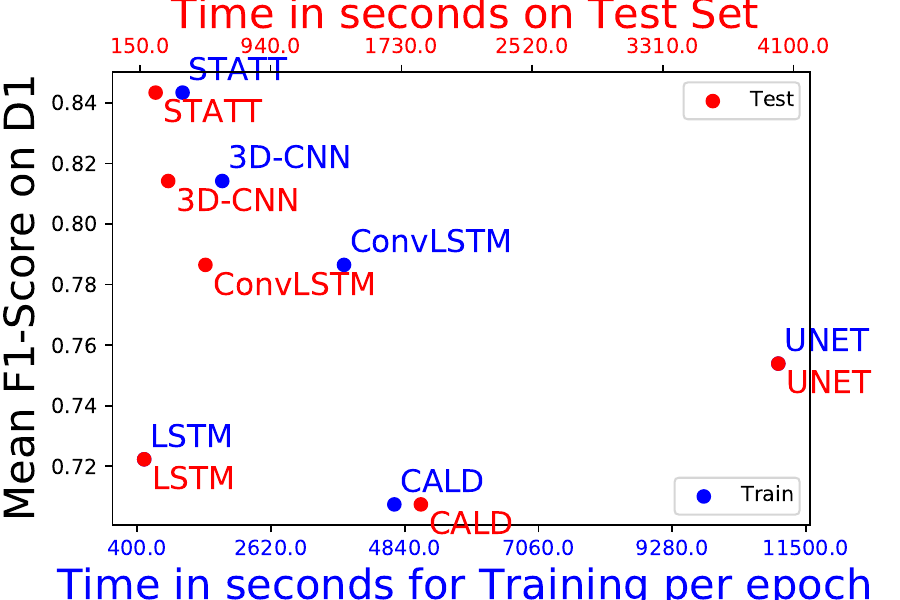}
    \caption{Mean F1-score vs Train and Test time for various methods}
    \label{fig:f1vstime}
\end{figure}

\subsection{Predictive Performance}

\begin{figure*}[t]
    \centering
    \includegraphics[width = \textwidth]{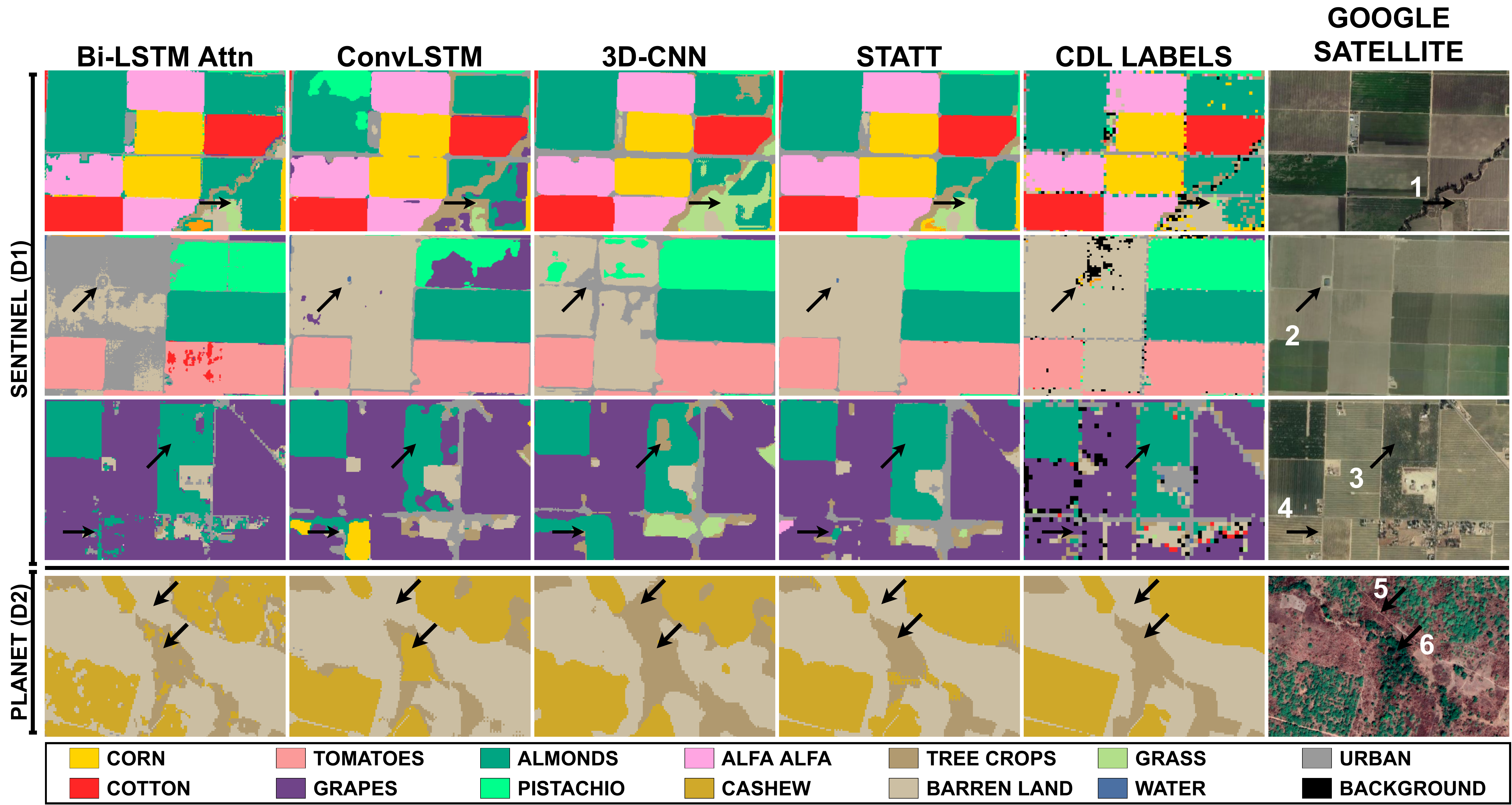}
    \caption{Predicted segmentation maps for baselines and STATT along with groundtruth labels from CDL and high resolution satellite image. Please zoom for better view}
    \label{fig:segmentationmaps}
\end{figure*}

The mean F1 score along with the class-wise F1 score for all the algorithms are reported in Table~\ref{Tab:tableResults}. We also report the scores for a variation of STATT, namely STMean, where instead of using attention to aggregate the spatio temporal featues we take the mean of the features and pass it to the decoder. We also report the number of pixels of each class along with their percentage of the total. We observe that STATT achieves the best  or the second best performance for most land cover classes (and the  best mean F1-score). 
It can also be seen that the incorporation of both spatial and temporal information leads to an increase in F1 score in both datasets. On one hand, UNet has worse performance than the spatio-temporal models because it uses only spatial data for prediction and lacks temporal modeling. On the other hand, the Bi-LSTM Attn approach performs the classification at the pixel level by focusing only on the temporal information, disregarding the spatial correlation. Hence, the predicted maps can contain more errors at individual pixels, resulting in degraded performance. 3D-CNN is one of such methods which exploit both the spatial and temporal information. Since the 3D-CNN model uses convolution in the time-dimension to model the temporal information, it does not track long-term changes but only considers a few temporally adjacent frames. ConvLSTM models the long-term temporal information using LSTMs but lacks due to its inability to model the discriminative time-periods for different land covers.

Moreover, ConvLSTM has a much larger number of parameters, which requires more training data and higher computational cost. Compared with these approaches, STATT models the spatial correlation and the temporal progression while identifying the discriminative time-periods. Moreover, STATT has much fewer parameters and thus can perform well even in the data scarcity scenario. The benefit of the attention mechanism can be sensed by comparing the performance of STMean and STATT. 

Figure \ref{fig:f1vstime} shows a scatterplot to demonstrate the computational efficiency of our approach, in which we compare Train and Test time on D1 along with the Mean F1-Score of the approach. It can be observed that \textit{ST-ATT} is able to obtain higher prediction performance (F1 score) at relative low computational cost among the spatio-temporal methods. \textit{ConvLSTM} tends to take about 10 times more time than \textit{ST-ATT}, which generally takes half the time taken by \textit{3D-CNN}. Note that the high runtime of UNet is due to the fact that the model need to be applied to each time step independently.

\begin{figure*}[t]
    \centering
    \includegraphics[width = \linewidth]{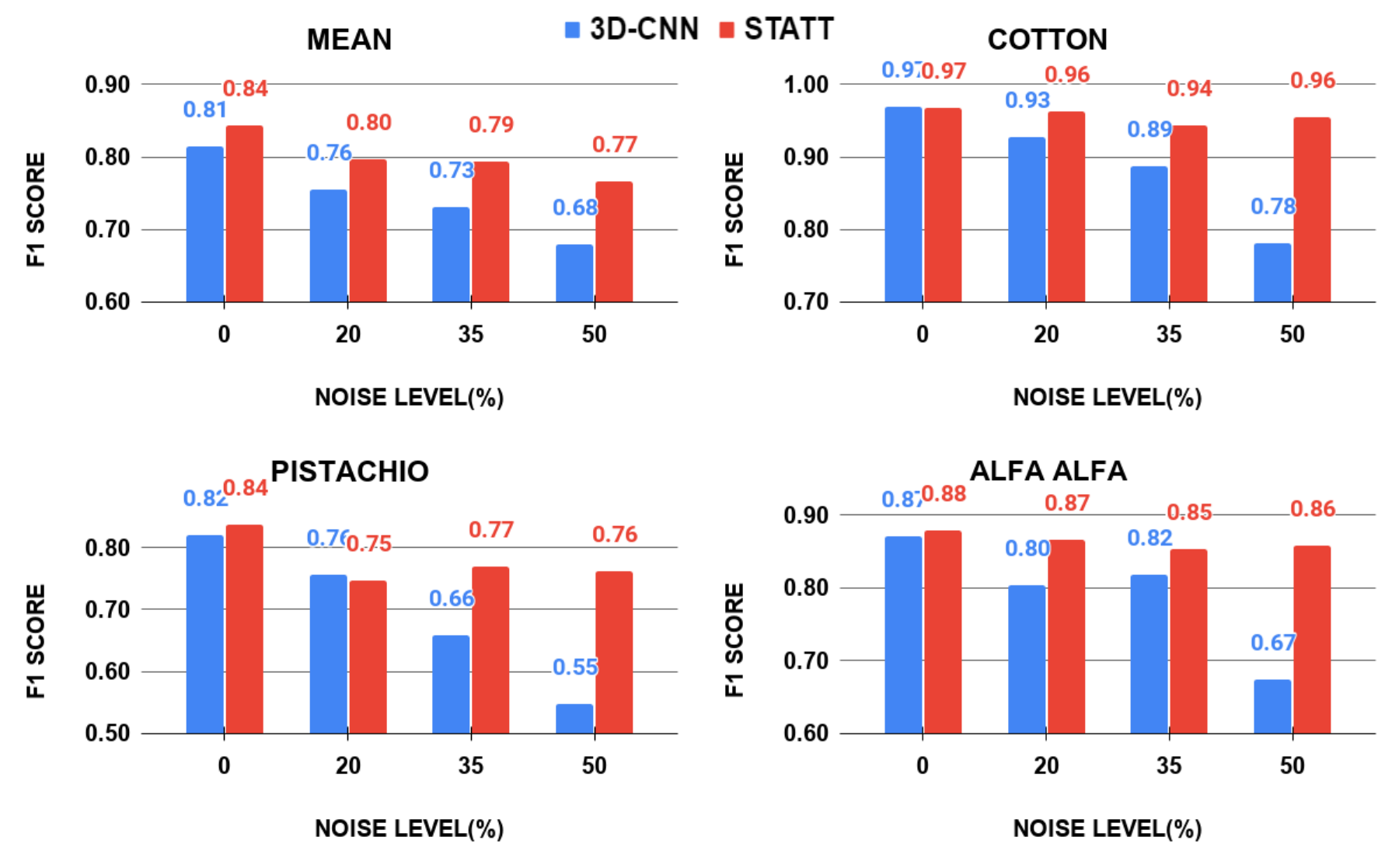}
    \caption{Impact of noisy images on the predictive performance for multiple land covers. The x-axis shows the percentage of noisy time-steps in the time-series and y axis shows F1-Score}
    \label{fig:noise_analysis}
\end{figure*}

\subsubsection{\textbf{Segmentation Results}}\hfill\\
Figure \ref{fig:segmentationmaps} shows predicted segmentation maps on randomly selected patches along with ground-truth labels from CDL and high resolution google satellite images. We highlight the important areas using arrows for easy reference. Row 1, 2 and 3 shows of example from the Sentinel dataset (D1). The first row consists of an area of land with various crop varieties. Bi-LSTM Attn results in a predicted map that is not spatially smooth which highlights the quality of maps produced by spatio-temporal methods over their pixel-based counterparts. Moreover, STATT is best able to produce a spatially continuous and correctly classified output, as can be observed by arrow 1. There are also errors in other methods such as ConvLSTM and 3D-CNN in the almond fields, which can be observed in the top part of the field. 

The second row shows another crop patch on which STATT produces the cleanest prediction map. The other methods suffer from false predictions of urban within the barren area. ConvLSTM manages to produce false urban but gives false grapes which can be observed in the top right corner of the field. In this field as well, one can observe that the prediction map of Bi-LSTM Attn is not continuous, as seen by speckles of Cotton within a field of Tomatoes. Another interesting detection of STATT is the fact that it identifies a water reservoir, marked by arrow 2, which is not shown even in the groundtruth USDA labels.

The third row shows a patch of grapes and almonds. As observed by arrows 3 and 4, STATT produces the best map, correctly distinguishing between classes like grapes, tree crops and almonds. We notice that ConvLSTM and 3D-CNN predict wrong crop classes at arrow 4 and produce noncontinuous maps at arrow 3.

Similarly, in the fourth row we show an example from the Planet dataset (D2). Due to lack of convolutional layers in CALD, the predicted map produced is not spatially smooth spatially smooth due to the lack of spatial contextual information. 3D-CNN and ConvLSTM produce maps which show confusion between Tree Crops and Cashew as denoted at arrows 5 and 6. From the qualitative analysis of the predicted maps, it can be seen that STATT not only detects challenging classes but also has less false positives when compared to the baselines. 

\begin{figure*}[t]
    \centering
    \includegraphics[width = \linewidth]{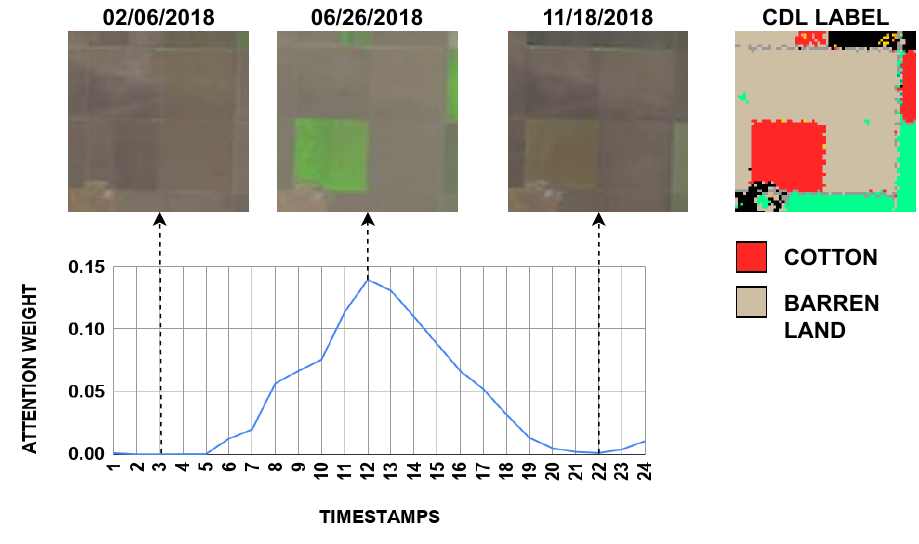}
    \caption{Visualisation of Attention Weights predicted for Cotton on Sentinel data.}
    \label{fig:attention_analysis}
\end{figure*}

\subsection{Robustness to Noisy Images}
STATT can reduce the effect of noisy time-steps by assigning low attention weights to the noisy time-steps. We demonstrate this empirically in Figure~\ref{fig:noise_analysis} by increasing the number of noisy images from 0\% to 50\%. In particular, we manipulate image sequence in D1 by replacing the clean images with Sentinel images containing noise (cloud and other atmospheric disturbances) from the same time-period. We report the change in the mean F1 score as well as the F1 score of an example class from each type, CROP(Cotton), TREE-CROPS(Pistachio) and COVER-CROP(Alfa-Alfa). From the results in Figure \ref{fig:noise_analysis}, we observe that the mean and class-wise F1 score of both STATT and 3D-CNN decreases with the increase in noise percentage. However, because of the use of attention based aggregation, the performance of STATT is only slightly affected compared to the drop in F1 score (both mean and class-wise) of 3D-CNN.

\subsection{Visualization of Attention Weights}
Here we visualize the obtained attention weights to verify the effectiveness of the attention mechanism in identifying the discriminative period specific to a land cover and thus helps distinguish between different land covers. In Figure \ref{fig:attention_analysis}, we plot the attention weights assigned to each time-step for the cotton class(in red). To verify the attention weights, we show the satellite image corresponding to three time-steps(3, 12 and 22). In the first image(02/06/2018), the crop is not yet planted and thus the farm is indistinguishable from a barren patch of land nearby which verifies the low attention weight. In the second satellite image(06/26/2018) it can be easily observed that the cotton is fully grown and thus it can be easily distinguished from a barren patch of land and thus the attention weight is the highest for this time-step. Finally, in the last time-step, the crop is harvested leaving the land barren which explains the low attention weight.

\section{STATT at work}
In Sec~\ref{Sec:Evaluation}, we demonstrated the performance of the proposed method STATT on test regions in the US and Benin, Africa. For the success of such deep learning methods in the task of RS-based crop mapping, existence of large-scale bench-marking datasets is essential. The availability of remotely-sensed satellite images provides tremendous opportunity to create such datasets. However, the availability of high-quality labels at desired temporal and spatial resolution still remains a challenge. Although, the United States Department of Agriculture (USDA) provides a publicly available land-cover classification map annually at 30m resolution for the conterminous Unites States (CONUS), there are several challenges highlighted in~\cite{reitsma2015does,calcrop21}. The proposed methods is a part of a generalized framework presented in~\cite{calcrop21} for creating such large-scale datasets at a desired year and temporal frequency. Specifically, in additions to other processing steps, STATT uses the spatial and temporal information to generate better quality labels using the noisy CDL as initial labels.

The dataset presented in~\cite{calcrop21} covers the entire California Central Valley Crop Belt of ~44,000 sq. km and provides pixel-wise labels covering 29 classes at a spatial resolution of 10m along with biweekly remotely-sensed images from the Sentinel-2 for the year 2018. Extensive validation of the crop-map product via quantitative and qualitative metrics using both manual as well as automated methods are provided to show the superiority of the labels over CDL. Such datasets are crucial for the generalization of deep learning methods and the success of deep learning in RS tasks such as land-cover mapping.

\section{Conclusion and Future Work}
\label{Sec:Conclusion}
Our proposed STATT architecture highlights the advantages of bringing in both spatial and temporal aspects of data for land cover mapping, especially in capturing the spatial correlation and temporal progression for each land cover. The experimental results on real world datasets have shown the superiority of STATT over its counterparts. Attention based mechanism helps in boosting the overall performance, especially in scenarios where the images are affected by clouds and other atmospheric disturbances. It also identifies the discriminative time-periods which is essential in identifying different land-covers. We believe that with these advances, such techniques have a huge potential in applications such as yield estimation, crop insurance, and improving management practices to maximize yield. The method is also a key component of the framework used in developing a crop-map product~\cite{calcrop21}, which is much more comprehensive than the state-of-the-art CDL labels in the diverse region of California Central Valley.

The proposed scheme can be extended along many directions. Traditional machine learning models treat each land cover separately but do not take into account the similarities amongst a large number of land cover classes. Land cover classes have a semantically meaningful hierarchy among them, e.g., corn and cotton are examples of crops, whereas peaches, grapes and almonds belong to the tree crop category. Standard practice of training neural networks via stochastic gradient descent do not account for the semantically meaningful organizations of the classes. Failure to capture such land cover relationships can raise two issues: 1) The distribution of different land covers can be highly biased and thus deep learning models can learn a poor representation for those land cover classes that appear less commonly in training data; 2) Given the hidden representation learned by the deep learning model, class confusion (misclassification) can happen between any classes that are close in the hidden space. However, in practice it may cause a more serious outcome if misclassification occurs across different high-level classes, e.g., crops vs. grassland, compared with misclassification within a high-level class, e.g., corn vs. soybean. Such hierarchical relationships can be encoded in the form of a topology graph. Specifically, we can add a regularization penalty in addition to the standard cross-entropy loss, where the aim is to make the embeddings (i.e. the output of the decoder) of two pixels to be similar if they belong to the same super-class. This can be done in the form of graph regularization, where we construct an output graph $G$ with the pixels as nodes. For two nodes $(i,j) \in V$, the edge weight $E(i,j)$ is higher if they belong to the same super-class, and thus we want the embeddings $(e_i, e_j)$ of these pixels to be similar as shown,
\begin{equation}
    \mathcal{L}_{Hier} = \frac{1}{|E|}\sum_{i,j \in V} E(i,j)\|e_i - e_j\|^2 = \frac{1}{2|E|}tr(\textbf{e}^T \text{Lap}(G) \textbf{e})
\end{equation}
where, $Lap(G)$ is the Laplacian matrix for the output graph $G$. For efficient computation and faster run-time, in each epoch we can randomly select $R$ pixels for each super-class to construct this output graph. The model is thus trained according to the objective:
\begin{equation}
    \mathcal{L} = \mathcal{L}_{CE} + \lambda \mathcal{L}_{Hier}
\end{equation}
where, $\lambda$ is a hyper-parameter to balance the cross-entropy loss and the hierarchical loss.

\section{Acknowledgments}
This work was funded by the NSF awards 1838159 and 1739191. Rahul Ghosh is supported by UMII MNDrive Graduate Fellowship. Access to computing facilities was provided by the Minnesota Supercomputing Institute.

\bibliographystyle{plain}
\bibliography{main_statt}

\vspace{12pt}

\end{document}